\documentclass[]{paper}

\usepackage{amsmath,amsfonts,bm}

\def\eqref#1{equation~\ref{#1}}

\def\1{\bm{1}}

\DeclareMathAlphabet{\mathsfit}{\encodingdefault}{\sfdefault}{m}{sl}
\SetMathAlphabet{\mathsfit}{bold}{\encodingdefault}{\sfdefault}{bx}{n}

\usepackage{float}
\usepackage{graphicx}
\usepackage{multirow}
\usepackage[utf8]{inputenc} 
\usepackage[T1]{fontenc}    
\usepackage{hyperref}       
\usepackage{url}            
\usepackage{booktabs}       
\usepackage{amsfonts}       
\usepackage{nicefrac}       
\usepackage{microtype}      
\usepackage{xcolor}         
\usepackage{amsmath}
\usepackage{epigraph}
\usepackage{graphicx}
\usepackage{colortbl}
\usepackage{tablefootnote}
\usepackage{longtable}

\usepackage{adjustbox}
\newcommand{\afflogo}[3][2.5ex]{%
  \raisebox{-0.35ex}{\includegraphics[height=#1]{#2}}%
  \hspace{0.25em}#3%
}

\usepackage{multirow}
\usepackage{xspace}
\usepackage{dsfont}
\usepackage{marvosym}

\newcommand{\authormeta}[1]{{\normalfont\mdseries\small #1}}
\newcommand{\authornote}[1]{{\normalfont\footnotesize #1}}
\newcommand{\corremail}{%
  {\color{metablue}\raisebox{-0.06ex}{\scriptsize\Letter}}\hspace{0.35em}%
}

\usepackage{listings}
\lstdefinelanguage{yaml}{
  keywords={true,false,null,y,n},
  keywordstyle=\color{blue!70!black}\bfseries,
  sensitive=false,
  comment=[l]{\#},
  commentstyle=\color{gray}\itshape,
  stringstyle=\color{purple!70!black},
  morestring=[b]',
  morestring=[b]",
}
\lstdefinestyle{nbcode}{
  basicstyle=\scriptsize\ttfamily,
  breaklines=true,
  breakatwhitespace=true,
  columns=fullflexible,
  keepspaces=true,
  showstringspaces=false,
  xleftmargin=4pt,
  frame=none,
  upquote=true,
  literate={\$}{{\$}}1,
}
\definecolor{positionblue}{RGB}{38, 86, 140}
\definecolor{positiongray}{RGB}{246, 248, 250}
\newcommand{\positionbox}[1]{%
  \begin{center}
  \setlength{\fboxsep}{7pt}%
  \colorbox{positiongray}{\parbox{0.92\linewidth}{#1}}%
  \end{center}
}

\definecolor{rccolor}{HTML}{FFF3E0}
\definecolor{tdcolor}{HTML}{E8F5E9}
\definecolor{iecolor}{HTML}{E3F2FD}
\definecolor{rcheader}{HTML}{E65100}
\definecolor{tdheader}{HTML}{2E7D32}
\definecolor{ieheader}{HTML}{1565C0}
\definecolor{headerrow}{HTML}{37474F}
 
\hypersetup{colorlinks=true, linkcolor=blue, urlcolor=blue!70!black}

\title{Interactive Evaluation Requires a Design Science}

\author{
Keyang Xuan$^{1,5*}$,
Peiyang Song$^{2,3*}$,
Pan Lu$^{4}$,
Pengrui Han$^{5}$,
Wenkai Li$^{3}$,
Zhenyu Zhang$^{4}$,
Zexue He$^{4}$,\\
Wenyue Hua$^{6}$,
Manling Li$^{7}$,
Jiaxuan You$^{5}$,
Adrian Weller$^{8}$,
Yizhong Wang$^{1\dagger}$,
Jiaxin Pei$^{1,4\dagger}$
\\[1.4em]
\authormeta{%
$^1$\afflogo[2.2ex]{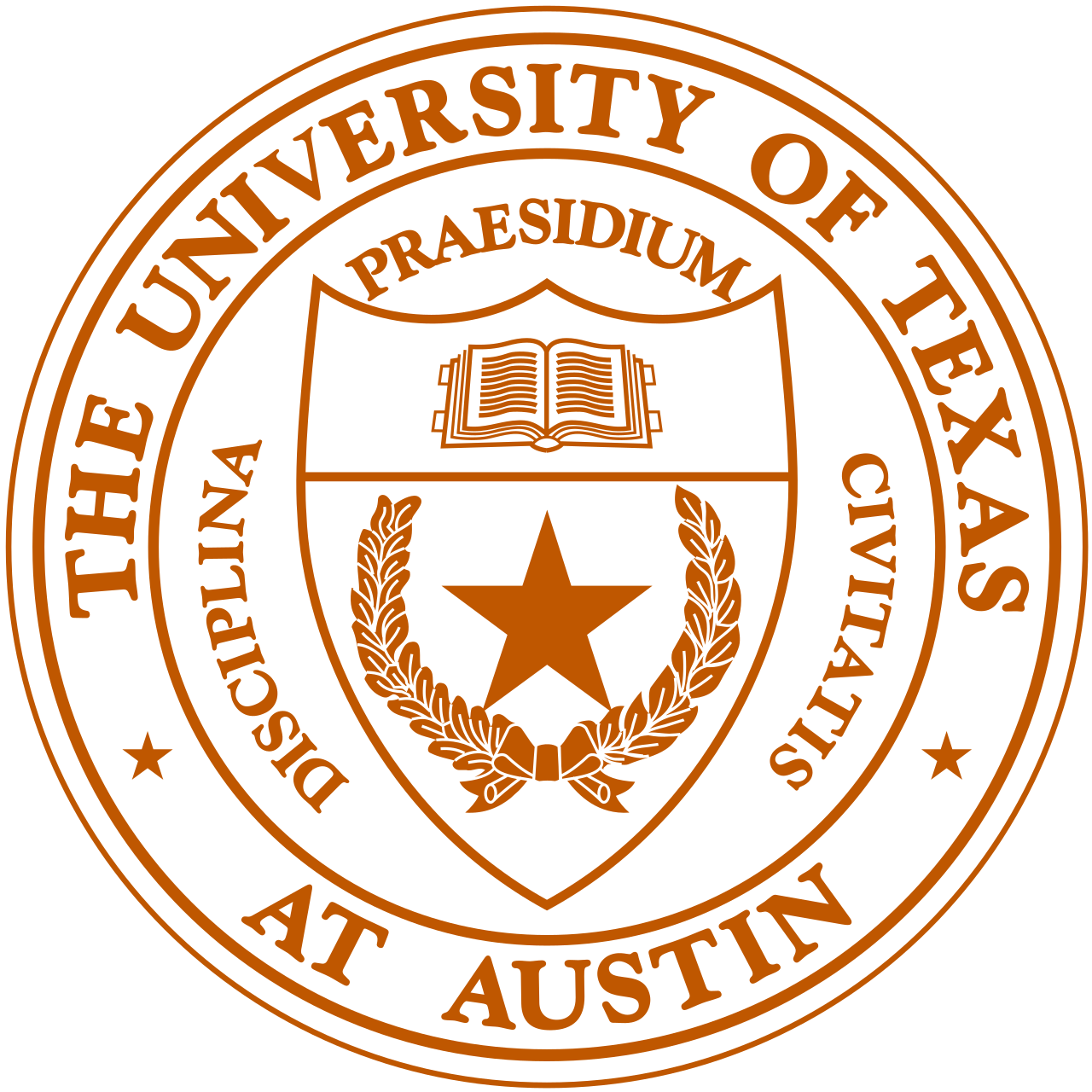}{University of Texas Austin}
\quad
$^2$\afflogo[2.2ex]{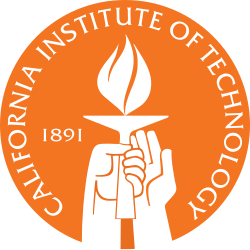}{California Institute of Technology}
\quad
$^3$\afflogo[2.2ex]{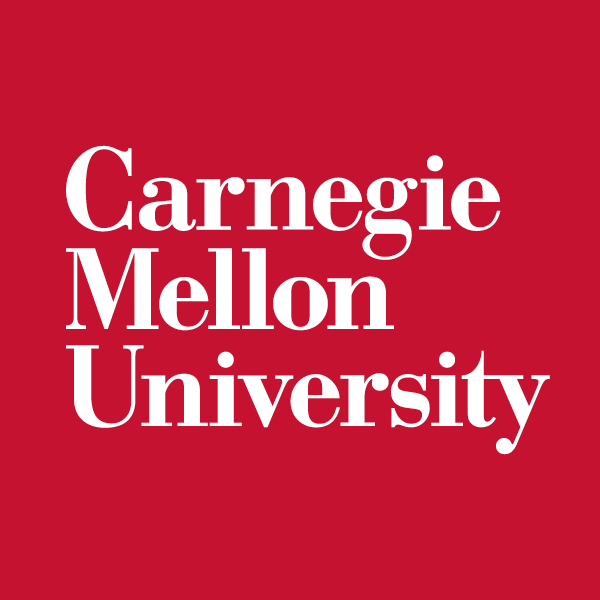}{Carnegie Mellon University}
\\[0.65em]
$^4$\afflogo[2.2ex]{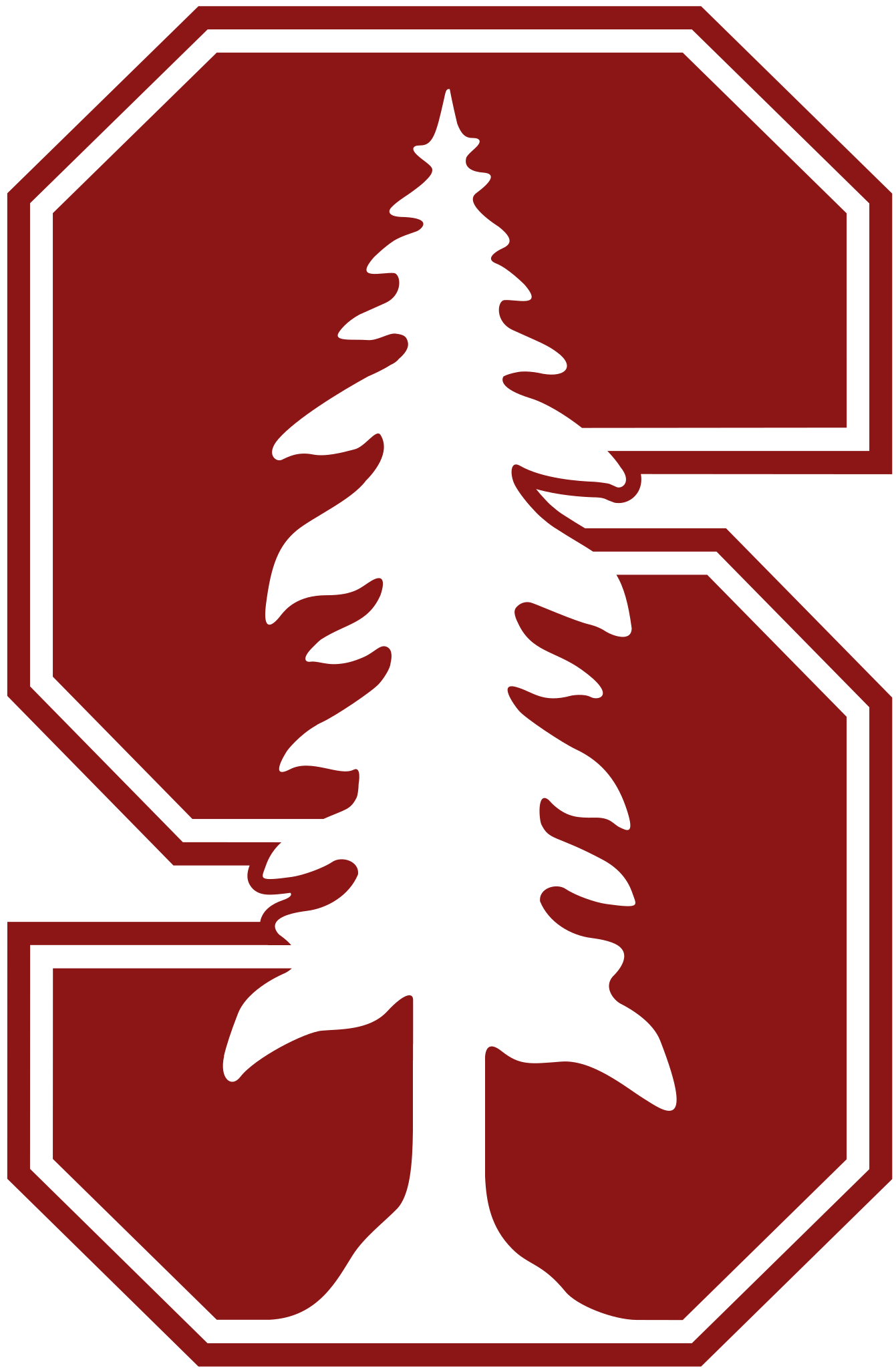}{Stanford University}
\quad
$^5$\afflogo[2.2ex]{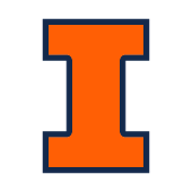}{University of Illinois Urbana-Champaign}
\quad
$^6$\afflogo[2.2ex]{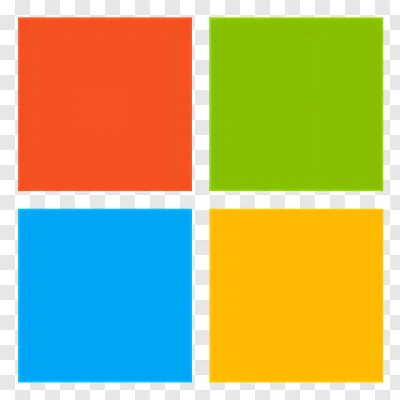}{Microsoft Research}
\\[0.65em]
$^7$\afflogo[2.2ex]{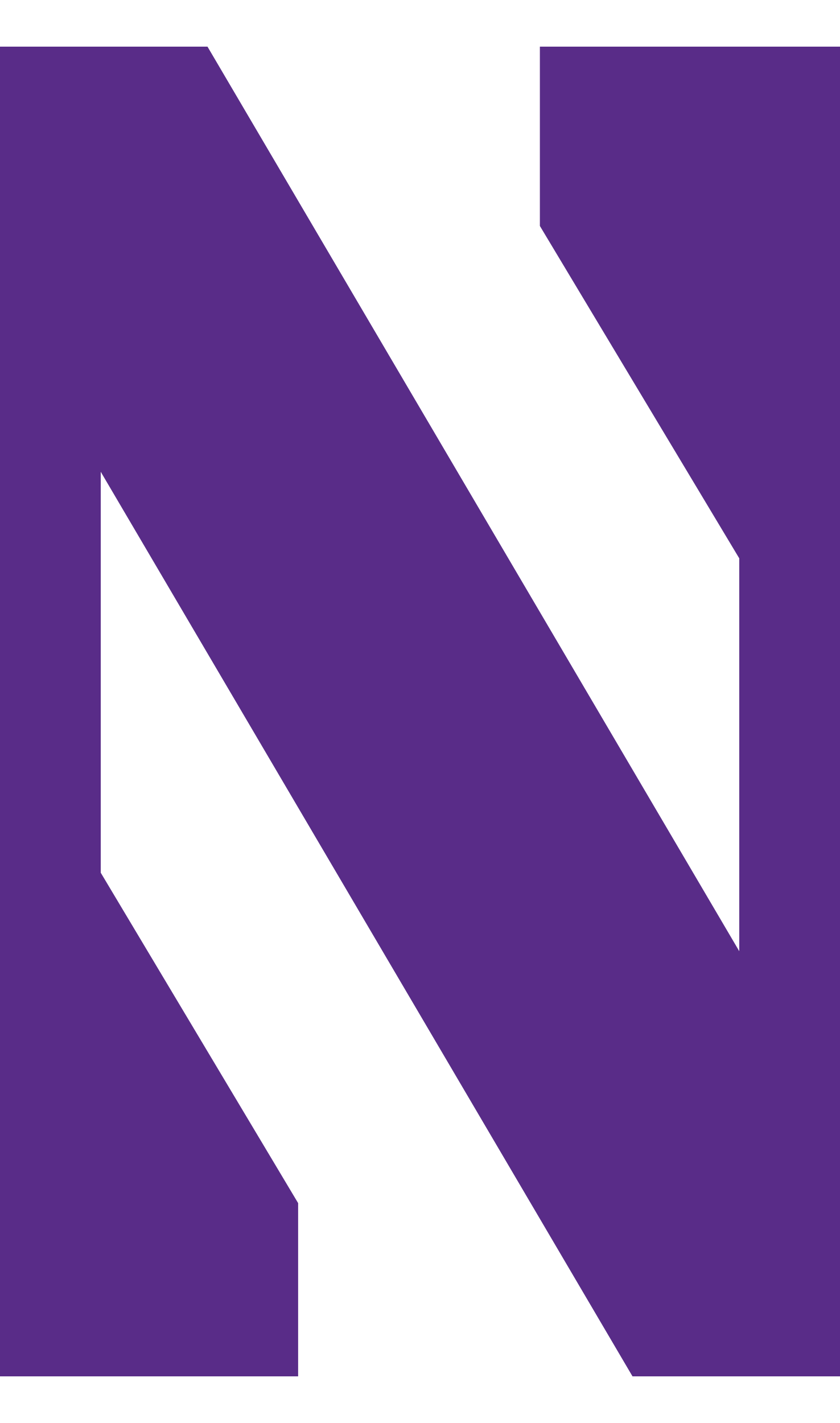}{Northwestern University}
\quad
$^8$\afflogo[2.2ex]{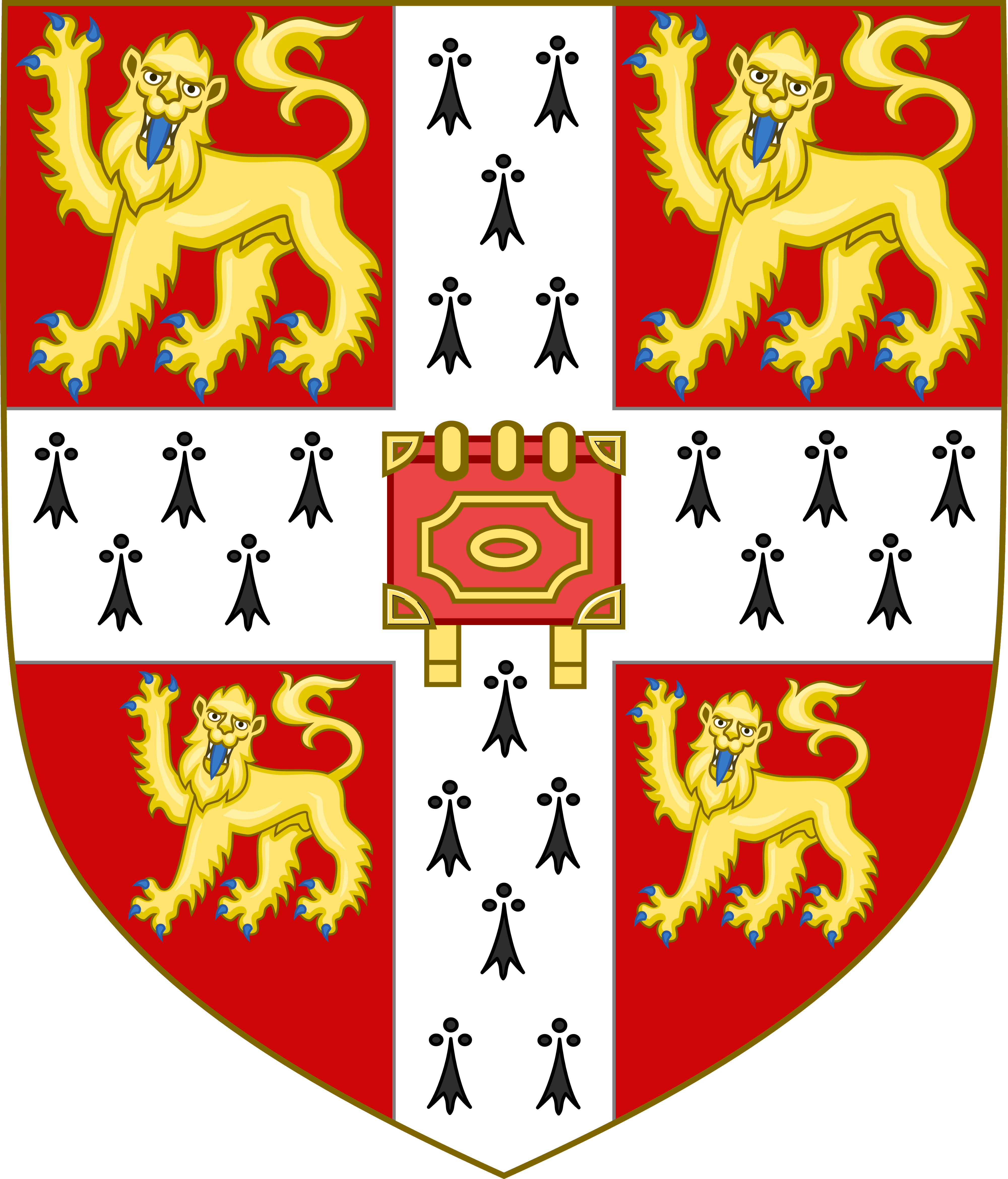}{University of Cambridge}
}%
\\[0.9em]
\authormeta{%
\corremail
\email{keyangx@utexas.edu}%
\,\textperiodcentered\,%
\email{psong2@andrew.cmu.edu}%
}%
\\[0.35em]
\authornote{$^{*}$Equal contribution\quad $^{\dagger}$Equal advising}%
}

\metadata[Code]{\url{https://github.com/keyangds/interactive_evaluation}}

\vspace{-2mm}
\abstract{
AI evaluation is undergoing a structural change. 
Large language models (LLMs) are increasingly deployed as systems that act over time through tools, environments, users, and other agents, yet many evaluation practices still inherit assumptions from response-centered benchmarks: fixed inputs, isolated outputs, and judgments made from a single response. 
Although interactive benchmarks have emerged, the landscape remains fragmented: benchmarks differ in what interaction artifacts they admit, how trajectories are scored, and what claims their results support. 
This position paper argues that interactive evaluation should be treated as a principled evaluation paradigm, not merely a new family of agent benchmarks. Simply adopting previous evaluation paradigms does not suffice. 
We define evaluation as an autonomous mapping from evidence to judgments, and show that interactive evaluation changes both sides of this mapping: the evidence becomes interaction-generated trajectories, while the evaluation procedure must assess process, recoverability, coordination, robustness, and system-level performance.
Building on this definition, we propose a two-axis taxonomy, derive design principles and reporting standards, examine representative scenarios, and analyze how longstanding evaluation challenges reappear at the trajectory level.
}

\begin{document}

\maketitle
\thispagestyle{plain}

\section{Introduction}
\label{Intro}
\vspace{-0.5mm}

\begin{figure*}[t]
  \centering
  \vspace{-10mm}
  \hspace{-4mm}
  \includegraphics[width=0.86\textwidth]{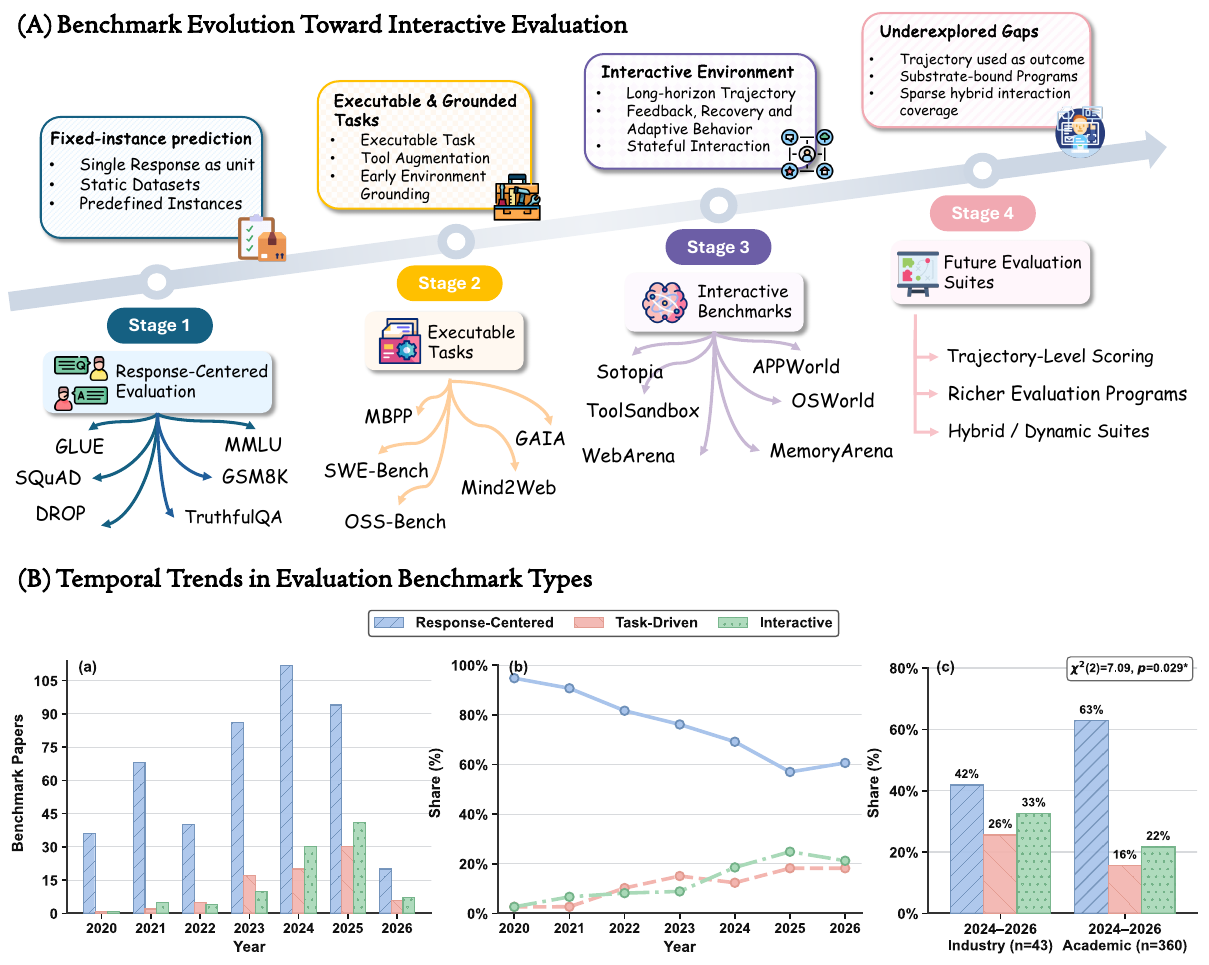}
\caption{
    \textbf{The rise of interactive evaluation motivates a design-science view.}
    Benchmarks have evolved from fixed-instance response evaluation to executable, grounded, and interactive settings, revealing gaps in trajectory-level scoring, substrate-independent programs, and evaluations combining tools, users or agents, memory, and changing environments.
    Empirical trends show growth in task-driven and interactive academic benchmarks, while frontier-lab reports show a more balanced evaluation-type mix.
    See Appendix~\ref{app:benchmark_collection} and~\ref{app:benchmark_list} for collection details and representative benchmarks. 
}
  \vspace{-3mm}
  \label{fig:roadmap}
\end{figure*}

AI evaluation is undergoing a visible transition. 
For much of modern AI, benchmark design was organized around response-centered evaluation: models received fixed instances and were judged by the quality of standalone final outputs, rather than by behavior unfolding through interaction. 
As Figure~\ref{fig:roadmap} illustrates, benchmark design has increasingly expanded toward executable, grounded, and interactive settings. 
This shift reflects a broader change in what large language models (LLMs) are expected to do: they are increasingly evaluated not only as standalone generators~\citep{gruver2023large, zheng2023judging, dubois2024length}, but as systems acting through tools~\citep{qin2023toolllm, schick2023toolformer}, interfaces~\citep{deng2023mind2web, patil2024gorilla, li2023api, zhang2023mobile, yao2022webshop}, environments~\citep{yao2022react, liu2024agentbench, shinn2023reflexion, shridhar2020alfworld}, external databases~\citep{karpas2022mrkl}, users~\citep{chalamalasetti2023clembench,lee2022evaluating}, and other agents~\citep{chen2023agentverse,li2023staticdatasetsdeepinteraction, jiang2025adaptation}. 
Across web navigation~\citep{zhou2023webarena}, tool use~\citep{guo2024stabletoolbench}, coding~\citep{jimenez2023swe}, formal mathematics~\citep{collins2025ai} and multi-agent coordination~\citep{emde2026maseval}, the object of evaluation is shifting from an isolated response to behavior that unfolds through feedback, state, and consequence~\citep{wang2023mint, xiagentgym, froger2025scaling, oktar2025identifying, song2024lean, Yang2024SWEagentAI}. 
This is not a cosmetic change in benchmark format. 
It changes what evidence an evaluation must observe and what claim a score can support.

The question is therefore no longer whether interactive evaluation ought to matter.
Recent benchmarks have already established its importance.
The urgent question instead is how interactive evaluation should be designed so that it becomes interpretable, comparable, and scientifically useful.

Existing interactive evaluations vary in the artifacts they record, the substrates and environments they include, the extent to which later states depend on earlier actions, and the procedures by which trajectories become scores. 
Some primarily test long-horizon goal completion in grounded environments~\citep{feng2026longcli}; others emphasize tool-user interaction~\citep{yao2024tau, lu2025toolsandbox, ibrahim2025towards}, process-level reward modeling~\citep{wang2026aligning}, social interaction~\citep{zhou2023sotopia}, or robustness under imperfect guidance~\citep{fu2026beyond}. 
These differences are productive, but they are also consequential to evaluation. 
A benchmark that records a trajectory but scores only final success supports a different claim from one that measures recoverability, risk, coordination, or adaptation. 
Without a shared conceptual frame, these distinctions are easy to flatten into a single category called ``agent evaluation,'' obscuring which evaluative claims are already well supported by existing benchmarks and which remain systematically under-covered.

This design problem is sharpened by an uneven transition across the evaluation ecosystem. 
As Figure~\ref{fig:roadmap} suggests, task-driven extensions and interactive evaluation appear more prominently in recent frontier-lab evaluation reports, while academic benchmark work still retains a stronger center of gravity around response-centered evaluation. 
We do not interpret this divergence as a simple gap in sophistication. 
Rather, it reflects different optimization pressures: academic benchmarks often prioritize  comparability, reproducibility, meaningful and scalable problem definition, while deployed systems increasingly require evidence about long-horizon interaction, tool use, robustness, and system behavior under feedback. 
As a result, different communities are beginning to optimize for different kinds of evidence and different kinds of evaluative claims. 
This makes it especially important to ask not only what trajectories a benchmark records, but also what evaluation program maps those trajectories to judgments.

\textbf{Therefore, this paper argues that:}

\positionbox{\textbf{Position.} Interactive evaluation should be built as a design science for evaluating systems acting through trajectories. The field does not merely need more interactive benchmarks; it needs explicit principles for specifying what interaction artifacts enter evaluation and how an evaluation program maps those artifacts to judgments.}

This paper develops that position from the perspective of evaluation itself. 
We first explain why response-centered evaluation was historically useful and why its assumptions become insufficient when systems act in closed loop. 
We then define evaluation as an autonomous program \(E: \mathcal{X} \rightarrow \mathcal{Y}\), where \(\mathcal{X}\) is the admissible evidence available to the evaluator and \(E\) is the procedure that maps that evidence to judgments. 
Interactive evaluation changes both parts: \(\mathcal{X}\) expands from final responses to interaction-generated trajectories, and \(E\) must assess not only final correctness but also process quality, recoverability, coordination, safety, efficiency, and robustness. 
This framing lets us build a taxonomy of interactive evaluation, use it to identify where current benchmarks concentrate and what they miss, and derive principles for designing future evaluations.

Our contributions are fourfold: 
\textbf{1)} We give a compact definition of interactive evaluation and clarify its boundary cases (Sec.~\ref{sec:definition}).
\textbf{2)} We propose a two-axis taxonomy organized around evaluation inputs and evaluation programs, making current and future benchmarks comparable without forcing them into one task domain (Sec.~\ref{sec:taxonomy}). 
\textbf{3)} We derive principles and a roadmap for benchmark design, reporting, and infrastructure (Sec.~\ref{sec:roadmap}). 
\textbf{4)} We illustrate the framework in representative coding-agent and multi-agent social-system scenarios (App.~\ref{app:illustrative_scenario}), then discuss risks that arise when classic evaluation problems--overfitting, gaming, leakage, brittleness, and reproducibility--become trajectory-level problems (Sec.~\ref{sec:risks}).

We therefore invite the community to treat interactive evaluation as a design science~\citep{simon2019sciences, hevner2008design, wieringa2014design}: one that specifies what trajectory artifacts count as evidence, how those artifacts are mapped to judgments, and what claims the resulting scores can support. The goal is not to evaluate harder tasks, but to evaluate interactive systems in ways that are interpretable, comparable, and scientifically useful.

\vspace{-2.5mm}
\section{Rethinking Evaluation Beyond Response-centered Evaluation}
\label{sec:rethinking}
\vspace{-1.5mm}

Response-centered evaluation is not a mistake to be discarded. 
It became dominant because it solved real methodological problems. 
Fixed datasets and standardized task instances made model comparison scalable; single-output scoring made results legible; and many core AI tasks could plausibly be represented as input-output mappings, including classification~\citep{bowman2015large, warstadt2019neural}, question answering~\citep{rajpurkar2016squad}, translation~\citep{goyal2022flores,tang2024creative,zhang2024hire}, summarization~\citep{gliwa2019samsum, kryscinski2022booksum}, and broad capability probing~\citep{hendrycks2020measuring, srivastava2023beyond}. 
In those settings, most relevant evidence is provided in the instance, the system's response is the natural unit of assessment, and later evaluation conditions do not depend on earlier model behavior.
\vspace{-2mm}
\paragraph{Why the Old Assumptions Worked.}

The response-centered paradigm matched a particular view of AI systems: a model receives an input \(x\), produces an output \(y\), and evaluation asks whether \(y\) has the desired relation to a reference, rubric, or judge. 
Its strength was not only convenience. 
It offered comparability, aggregation, and repeatability. 
These are still essential values. 
Interactive evaluation should supplement response-centered evaluation where interaction is constitutive of the capability being measured; it should not turn every evaluation into an expensive simulation by default.
\vspace{-1.75mm}
\paragraph{Why Interaction Breaks the Fit.}

The fit breaks when the system being evaluated acts over time. 
A web action can reveal or hide future opportunities; a tool call can modify persistent state; a user reply can change after clarification; another agent can adapt strategically; and an error can become recoverable rather than terminal. 
In these cases the evidence needed for judgment is not contained in the initial prompt or the final answer. 
It is generated through the trajectory. 
Several latest benchmarks~\citep{zhou2023webarena, xie2024osworld, trivedi2024appworld, wang2023mint, lu2025toolsandbox} make this visible by requiring systems to operate through executable environments, tools, or conversational feedback.
\vspace{-1.75mm}
\paragraph{Why Interaction Itself Must Be Evaluated.}

Interaction is not merely a path toward an answer; it is often the capability of interest. 
A coding agent that passes tests by making a brittle, unreviewable patch has not demonstrated the same competence as one that isolates the fault, preserves interfaces, and recovers from failing tests. 
A social agent that achieves a local objective by confusing a counterpart has not demonstrated the same competence as one that coordinates transparently. 
Once process changes the meaning of success, outcome-only measurement becomes under-specified. 
Interactive evaluation must therefore ask how evidence was gathered, which actions changed the state, whether mistakes were detected, and what costs or risks were incurred along the way.
\vspace{-1.75mm}
\paragraph{A Minimal Notion of Interaction.}

We use \emph{interaction} in a consequential sense. 
A setting is interactive when the system operates in an external loop involving tools, environments, users, or other agents; when what it encounters next depends at least partly on earlier behavior; and when that dependence matters for evaluation. 
Multiple turns alone are insufficient. 
A scripted dialogue whose later prompts are fixed in advance may be sequential, but it is not interactive in the evaluative sense used here. 
Conversely, a short tool-use task may be interactive if the tool result changes the subsequent evidence, state, or scoring conditions.

\section{Definition and Scope of Interactive Evaluation}
\label{sec:definition}
\vspace{-1mm}
An evaluation can be viewed as an autonomous program
\[
E: \mathcal{X} \rightarrow \mathcal{Y},
\]
where \(\mathcal{X}\) is the domain of artifacts accepted as evidence and \(\mathcal{Y}\) is the space of evaluative outputs, such as scores, rankings, pass/fail decisions, diagnostic reports, or qualitative judgments. 
This framing is intentionally simple but inevitable. 
Any scalable evaluation must decide what artifacts can be submitted to the evaluator and what procedure maps those artifacts to claims.

\positionbox{\textbf{Definition.} Interactive evaluation is evaluation in which the admissible evidence \(\mathcal{X}\) includes trajectories generated by consequential interaction, and the evaluation program \(E\) maps those trajectories to judgments about system-level performance.}


\begin{figure*}[t]
  \centering
  \hspace{-8mm}
  \includegraphics[width=0.85\textwidth]{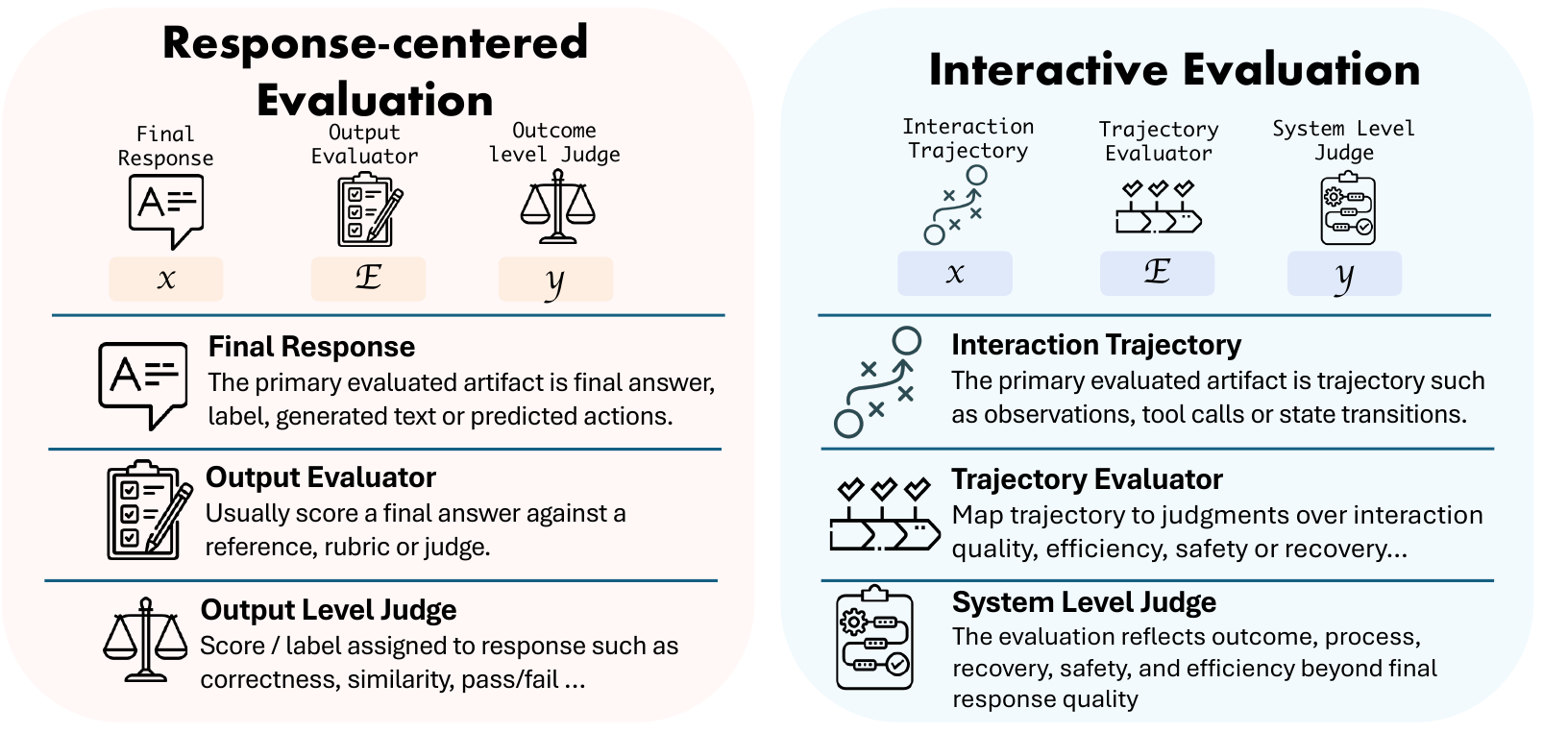}
  \caption{Response-centered evaluation judges final responses; interactive evaluation judges trajectories as evidence of system-level performance and process.}
  \vspace{-2.5mm}
  \label{fig:comparsion}
\end{figure*}

\vspace{-2mm}
\paragraph{What Changes in \(\mathcal{X}\).}

In response-centered evaluation, the central artifact is often a final answer, label, generated text, or predicted action for a predefined instance. 
In interactive evaluation, the artifact is a trajectory: observations, actions, tool calls, state transitions, user or counterpart responses, intermediate artifacts, costs, constraints, and final outcomes. 
The trajectory may come from a web environment~\citep{zhou2023webarena,he2024webvoyager}, an operating system~\citep{xie2024osworld}, a set of stateful apps~\citep{trivedi2024appworld}, a tool-user simulation~\citep{yao2024tau, lu2025toolsandbox}, or a social/multi-agent world~\citep{zhou2023sotopia, zhu2025multiagentbench}. 
What matters is not the substrate alone, but whether the recorded artifact preserves the action-dependent structure needed to judge performance.
\vspace{-1mm}
\paragraph{What Changes in \(E\).}

The evaluator also changes. 
A response-centered evaluator can often score a final answer against a reference, rubric~\citep{hashemi2024llm}, or autonomous judge~\citep{gu2024survey}. 
An interactive evaluator must decide which trajectory properties count: completion, progress, constraint satisfaction, efficient exploration, safe tool use, recoverability after error, cooperation, communication quality, or resilience under disruption. 
Thus \(E\) is not merely an answer checker; it is a trajectory-to-judgment procedure. 
It may combine executable tests, state checks, human or model judges, process annotations, penalties for unsafe actions, and aggregation across stochastic runs.
\vspace{-1mm}
\paragraph{Boundary Cases.}

Boundary cases are helpful in positioning the scope precisely.
This definition excludes three common false positives. 
First, multiple turns are not enough if the sequence is predetermined and earlier behavior does not affect later conditions. 
Second, tool calls are not enough if they are only hidden computation and do not change the evaluation evidence or state. 
Third, chain-of-thought or self-reflection is not enough by itself: internal reasoning may be valuable evidence when exposed under a protocol, but interaction requires an external loop whose continuation is partly action-dependent. 
The boundary is therefore evaluative rather than stylistic: a setting counts when judging the system requires evidence from consequential interaction.

\section{Taxonomy of Interactive Evaluation}
\label{sec:taxonomy}
\vspace{-1.5mm}
The definition above suggests that the main design problem in interactive evaluation is not simply whether a benchmark contains interaction, but whether its trajectory evidence is matched to an appropriate evaluation program.
We therefore use the evaluation mapping \(E: \mathcal{X} \rightarrow \mathcal{Y}\) as a diagnostic framework. Then interactive evaluations differ along two axes: what interaction-generated artifacts enter \(\mathcal{X}\), and how \(E\) maps those artifacts to judgments. 
This two-axis view avoids a common confusion: task domain, substrate, metric, and judgment protocol are not separate top-level taxonomies. 
They are properties of either the input artifact or the evaluation program.

\vspace{-1.5mm}
\subsection{Axis 1: Evaluation Inputs}

The first axis asks what interactive artifact is passed into evaluation. 
The central object is a trajectory, but trajectories differ in what they connect the system to.

\vspace{-1.5mm}
\paragraph{Tools and Environments.}

Many current benchmarks evaluate agents in executable digital or tool-mediated settings. 
WebArena, Mind2Web, BrowseComp, OSWorld, AndroidWorld, AppWorld, and MineDojo test interaction with web pages, interfaces, operating systems, mobile environments, stateful applications, or games~\citep{deng2023mind2web, zhou2023webarena, wei2025browsecomp, xie2024osworld, rawles2024androidworld, trivedi2024appworld, fan2022minedojo}. 
These inputs expose action-dependent state: clicks, API calls, file edits, or app operations change what the agent can observe later.

\vspace{-1.5mm}
\paragraph{Users.}

User-centered trajectories evaluate whether systems can interact effectively with people under incomplete, ambiguous, or evolving instructions. 
These evaluations focus not only on task completion, but also on whether systems can clarify user intent, maintain alignment with human goals, communicate uncertainty appropriately, and adapt as user preferences or requirements change over time. 
\(\tau\)-bench, IN3, ToolSandbox, MINT, RealWebAssist, and AgentClinic represent this direction by making user feedback or simulated user behavior part of the trajectory~\citep{yao2024tau, qian2024tellmoreimplicituser, lu2025toolsandbox, ye2026realwebassist, schmidgall2024agentclinic}. 
The artifact is therefore not just a task log; it includes how the system negotiates information, uncertainty, and coordination with users throughout the interaction.

\vspace{-2mm}
\paragraph{Other Agents.}

Multi-agent trajectories evaluate coordination, competition, delegation, negotiation, and emergent behavior. 
SOTOPIA, MultiAgentBench, BattleAgentBench, Intellagent, MASEval, and CooperBench show that the relevant evidence may include messages, role assignments, joint plans, conflicts, and counterpart adaptation~\citep{zhou2023sotopia, zhu2025multiagentbench, wang2024battleagentbench, levi2025intellagent, emde2026maseval, khatua2026cooperbench}.

\vspace{-1.5mm}
\paragraph{Hybrid and Dynamic Systems.}

The most deployment-like evaluations will combine tools, users, agents, memory, and changing environments. MemoryArena~\citep{he2026memoryarena} and large environment suites like AI Gamestore~\citep{ying2026ai} and ARC-AGI-3~\citep{foundation2026arcagi3newchallengefrontier} point toward persistent state and cross-session dependencies~\citep{froger2025scaling, backlund2025vending}. 
This category remains comparatively underexplored, but it is central for evaluating systems that must remain reliable across time rather than only within a single task episode, and we anticipate that this direction will soon become vital to a wide range of real-world tasks.

\vspace{-2mm}
\subsection{Axis 2: Evaluation Programs}
\vspace{-1mm}
The second axis asks how trajectories are mapped to judgments. 
Several measurement logics recur, and strong benchmarks should state which ones they support.

\vspace{-2mm}
\paragraph{Task Success.}

The most common program checks whether the final state satisfies a goal: a web task completed, a repository issue resolved, a mobile task performed, or an app state updated~\citep{zhou2023webarena, jimenez2023swe, rawles2024androidworld, trivedi2024appworld}. 
This is indispensable, but insufficient when two trajectories reach the same final state through very different risks or costs. This is the base-case evaluation where most principles from response-centered evaluation transfer, and will be supplemented by the interactive evaluation measures below.

\vspace{-2pt}
\paragraph{Process Quality and Efficiency.}

Interactive settings make intermediate behavior evaluable. 
A benchmark may score tool choice, action economy, state exploration, code-edit locality, communication clarity, or unnecessary disruption~\citep{wang2023mint, lu2025toolsandbox, yue2026interactive, Li2026ToolPRMBenchEA, george2025leanprogress, Fan2026AgentProcessBenchDS, bai2025and}. 
These measures are important because poor processes often predict brittle deployment even when final success is achieved.

\paragraph{Recoverability and Robustness.}

A trajectory-level evaluator can test whether systems detect mistakes, revise plans, resist misleading guidance, and remain effective under changing conditions~\citep{debenedetti2024agentdojo, fu2026beyond, froger2025scaling, han2025personality}. 
This is one of the clearest advantages of interactive evaluation: failure is not merely an endpoint, but an event that can be observed, repaired, or amplified.

\paragraph{Safety, Alignment, and Social Competence.}

When systems interact with users or agents, evaluation must include norm-sensitive behavior, cooperation, honesty about uncertainty, and avoidance of manipulative or unsafe strategies~\citep{zhou2023sotopia, khatua2026cooperbench, zhang2024agent, zhou2024haicosystem, song2026large, cemri2026multi, zhu2025llm}. 
These properties are often invisible in final-answer scoring but central to whether an interactive system should be trusted.

\positionbox{\textbf{Taxonomy Claim.} A benchmark is a point, or a region, in a two-dimensional design space: the interaction artifact it admits as evidence, and the trajectory-to-judgment program it implements. This view lets us compare benchmarks without pretending that all interactive tasks measure the same capability.}

\begin{figure*}[t]
  \centering
  \vspace{-4mm}
  \includegraphics[width=\textwidth]{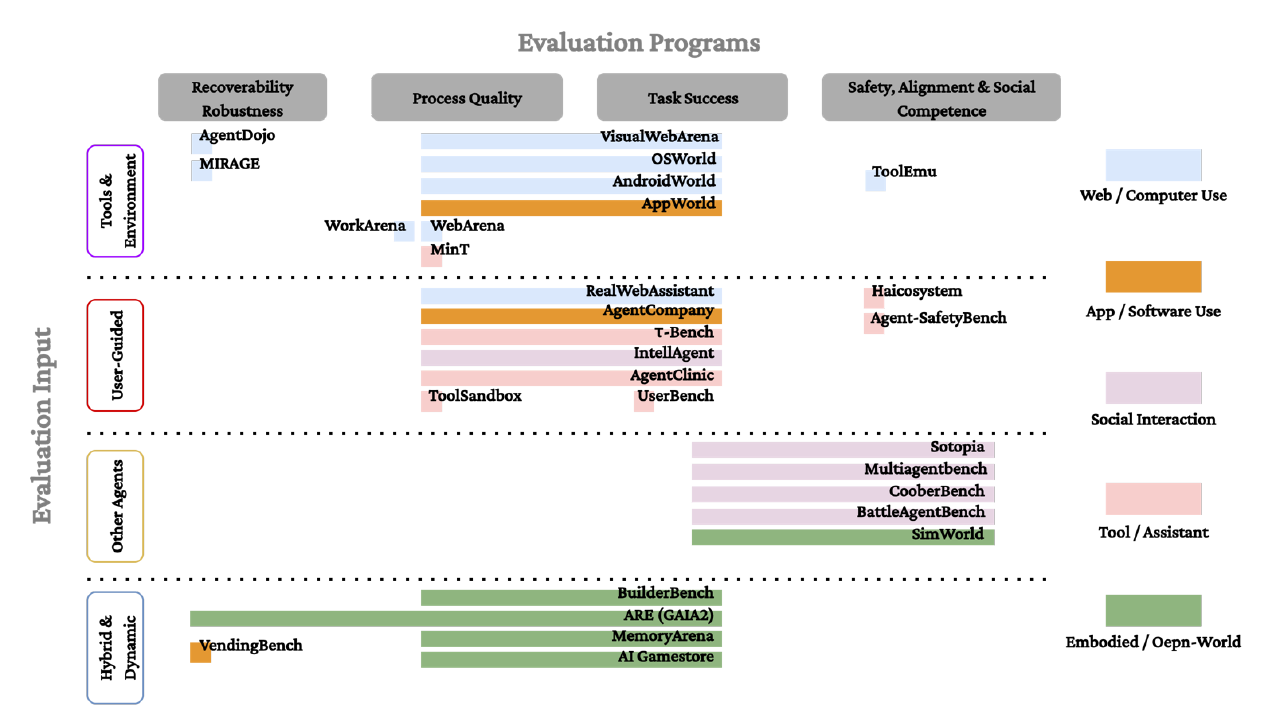}
  \caption{Mapping the current interactive evaluation landscape in a two-dimensional design space.}
  \label{fig:taxonomy}
  \vspace{-2mm}
\end{figure*}

\vspace{-2mm}
\subsection{Putting It Together: The 2D taxonomy \& Derived Observations}

Figure~\ref{fig:taxonomy} maps a representative set of interactive evaluations into the 2D space defined above. 
We do not claim to provide an exhaustive census; instead, we prioritize works with academic impact, open-source adoption, and use in frontier-model evaluation, drawing from the broader benchmark pool summarized in Appendix~\ref{app:benchmark_list}. 
This mapping provides a global view of existing interactive evaluation while highlighting underexplored areas. 
From this mapping, we derive the following observations:

\vspace{-2mm}
\paragraph{Trajectory Evidence Remains Outcome-Centered.} The most visible pattern is the concentration around Task Success and the relative sparsity of Recoverability and Robustness. This concentration indicates a mismatch between trajectory evidence and evaluation programs. Many works admit trajectories as evidence, but still evaluate them as final outcomes. As a result, interactive evaluation has often adopted trajectory recording without fully developing trajectory-level judgment. 
When trajectories contain actions, observations, state changes, and feedback, evaluation should also support judgments about process quality, cost, risk, and recoverability, rather than treating them only as evidence for a single success label.

\vspace{-2mm}
\paragraph{Evaluation Programs Remain Substrate-Bound.}

Many interactive evaluations remain organized around evaluation substrates rather than evaluation programs.
Tools and Environments are usually evaluated through Task Success or Process Efficiency; Other Agents are more often evaluated through Safety, Alignment, and Social Competence; and Hybrid and Dynamic Systems remain sparse across programs. 
This suggests that evaluation programs often follow what is easiest or most natural to measure in a given substrate, rather than beginning from claims such as recoverability under disruption, safety under persistent tool use, or robustness under changing user intent, and then designing the input artifact, protocol, and scoring procedure needed to support them.

\paragraph{Hybrid and Dynamic Systems Lack Robust Coverage.} As LLM systems face more complex end-to-end tasks, evaluation must increasingly account for hybrid and dynamic interaction evidence. Yet our mapping shows that this category remains the sparest evaluation input category. Existing work in this region remains concentrated around Task Success and Process Quality and Efficiency, leaving recoverability, safety, alignment, and social competence comparatively underdeveloped. This creates a mismatch between current evaluation coverage and likely deployment risk. 
In systems with persistent state, cross-session dependencies, and hybrid interaction loops, errors may accumulate over time instead of ending with a single failed task. As systems move toward longer horizons and more persistent operation, interactive evaluation must make these conditions first-class objects of measurement rather than treating them as future extensions.

\subsection{Why Current Interactive Evaluation Remains Underspecified}

The taxonomy above reveals global gaps in the current interactive evaluation landscape. These gaps matter because they also appear as implementation-level limitations in current evaluation practice. 
Many interactive evaluations are difficult to interpret, compare, audit, and build on because key design choices remain only partially specified. 
We highlight three recurring limitations in current practice.

\paragraph{Interactive Evaluations Are Costly to Run.}
Unlike traditional evaluation, interactive evaluations often require long-horizon rollouts, stateful environments, repeated runs, human or simulated counterparts, trajectory storage, and trajectory-level judging. 
These requirements make evaluation expensive and difficult to audit~\citep{ndzomga2026efficient}, especially for smaller research groups or independent evaluators. 

\paragraph{Human Validation Is Often Underused.}
Many interactive evaluations rely on final success scores, automated checks, or LLM-as-a-judge pipelines, even when the evaluated behavior involves user intent, social appropriateness, or domains that require human expertise. 
For these claims, human-in-the-loop validation is often needed. 
Without such validation, scalable scores may appear reliable while failing to capture user-facing failures that only become visible in context.

\paragraph{Execution Protocols Are Difficult to Reproduce.}
Interactive scores depend on concrete protocol choices, including tool access, observation space, retry budget, reset condition, or environment version. 
Small changes in these choices can alter what opportunities the system has, what failures mean, and whether two trajectories are comparable~\citep{mustahsan2025stochasticity, zhu2026establishing}. 
Without carefully designed protocol reporting, later work may appear to evaluate on the same benchmark while actually measuring different interaction settings.

\section{Principles and Roadmap for Interactive Evaluation}
\label{sec:roadmap}

The taxonomy and implementation-level diagnosis above show that interactive evaluation requires more than recording trajectories. 
Benchmarks must specify 
what system is evaluated, what trajectory evidence counts, what protocol generated it, which dimensions are judged, and what claims the scores support. 
These choices make benchmarks interpretable and comparable, and define what 
the field should build next.

\vspace{-2pt}
\paragraph{Specify the System and Trajectory Evidence.}

Interactive evaluations should specify the evaluated system, its accessible resources, the recorded trajectory evidence, and the claims that evidence supports. Model identity alone is insufficient: tool wrappers, memory, retrieval, execution sandboxes, user simulators, and orchestration policies can all affect performance. Recording trajectories is also insufficient unless benchmarks state whether those traces support judgments about task success, process quality, recoverability, safety, efficiency, coordination, or other system-level properties.

\vspace{-2pt}
\paragraph{Specify the Interaction Protocol.}

Trajectory evidence is only interpretable relative to the protocol that generates it. 
Evaluation should therefore specify the conditions under which interaction unfolds, including the initial state distribution, allowed actions, observation space, counterpart behavior, stopping rules, randomness, persistence, and reset conditions. 
These details determine what opportunities the system had, what constraints it faced, and whether two trajectories are meaningfully comparable. 
Protocol documentation is the interactive analogue of dataset documentation: without it, scores may reflect hidden differences in interaction setup rather than differences in system capability.

\vspace{-2pt}
\paragraph{Design for Perturbation and Repair.}

As evaluation tasks move toward more complex and dynamic environments, clean task completion becomes an insufficient test of interactive competence. Future benchmarks should therefore evaluate whether systems can remain effective when interaction conditions change, including ambiguity, misleading feedback, partial failure, state drift, and counterpart adaptation. These conditions should not be treated as adversarial add-ons; they are central to interactive evaluation because they reveal whether a system can detect problems, revise its strategy, recover from errors, and continue acting effectively under evolving conditions.

\vspace{-2pt}
\paragraph{Separate Outcome, Process, and Risk.}

Interactive evaluations should distinguish what the system ultimately achieves from how it achieves it and what risks it creates along the way. A single scalar score may be useful for ranking systems, but it should not hide distinct evaluative claims. Interactive benchmarks should therefore report final success separately from trajectory-level properties such as action cost, unsafe behavior, recovery behavior whenever those dimensions matter. Aggregated scores can still be useful, but they should be treated as summaries of multiple reported dimensions rather than as the only evidence of system capability.

\vspace{-2pt}
\paragraph{Build Shared Infrastructure without Freezing the Design Space.}

The field needs reusable environments, logging schemas, trajectory viewers, evaluation harnesses, and reporting templates. 
At the same time, standardization should preserve diversity in domains and protocols. 
A healthy roadmap moves from response-centered benchmarks, through executable and tool-augmented tasks, toward interactive suites that make protocol, state, and measurement logic first-class objects.

\section{Broad Risks and Open Issues in Interactive Evaluation}
\label{sec:risks}
The principles above address current gaps in interactive evaluation design. 
As the field moves toward better-specified evaluations for increasingly complex real-world tasks, a second class of risks comes into view. 
These are not merely failures of design, but risks that emerge as interactive evaluation becomes more capable, standardized, and consequential as shared scientific infrastructure. 
We therefore distinguish two classes of issues: longstanding evaluation risks that reappear at the trajectory level, and risks that become salient as consequential interaction becomes part of the measurement process.

\subsection{Longstanding Evaluation Risks at the Trajectory Level}

\paragraph{Overfitting, Leakage, and Gaming.}

Static benchmarks can be memorized; interactive benchmarks can be policy-gamed. 
In response-centered evaluation, leakage often concerns exposure to test inputs or reference answers. 
In interactive evaluation, leakage can occur through environment state, public task templates, tool APIs, simulator regularities, predictable user models, or evaluator heuristics. 
The resulting failure mode is not only that a model knows the answer, but that a system learns how to behave strategically inside the benchmark. 
Agents may exploit simulator quirks, avoid meaningful exploration, optimize for superficial trajectory signals, or discover shortcuts that satisfy the scorer without demonstrating the intended competence. 
Mitigations should therefore operate at the trajectory level: held-out environments, procedurally generated tasks, private or refreshed evaluation suites, adversarial perturbations, and audits of suspiciously efficient or unnatural trajectories.

\paragraph{Distribution Shift and Benchmark Brittleness.}

Benchmarks have risked measuring performance under a narrow data distribution. 
Interactive evaluation makes this problem more acute because small changes in interface, timing, initial state, tool behavior, or counterpart response can alter the trajectory itself. 
This sensitivity should not be dismissed as mere noise: deployment also involves shifting states, imperfect instructions, and changing counterparts. 
However, benchmarks must distinguish robustness failures from protocol artifacts. 
Reporting should include variance across seeds, environments, users, perturbations, and state initializations, and should identify whether failures reflect missing capability, brittle policies, or genuine sensitivity to deployment-relevant variation.



\subsection{Risks Native to Interaction}

\paragraph{Standardization--Diversity Tradeoff.}
Interactive evaluation should develop shared infrastructure without collapsing into a single narrow evaluation format. Shared logging schemas, reporting standards, and replay infrastructure are necessary for comparability, reproducibility, and auditability. 
At the same time, the field faces a second risk: premature convergence on a small set of protocols or environments. 
Excessive standardization can narrow the range of interaction patterns through which competence is defined. 
A useful ecosystem should standardize how evaluative claims are specified and reported while preserving diversity in environments, interaction substrates, counterpart models, and trajectory-level measurement programs.

\vspace{-4pt}
\paragraph{Fidelity, Control, and Simulator Artifacts.}

Interactive evaluation must decide how much of deployment to reproduce and how much to abstract away. 
High-fidelity environments can provide richer evidence about situated behavior, but they are expensive, noisy, and harder to control. 
Controlled simulators improve repeatability and comparison, but may reward strategies that exploit simulator artifacts rather than genuine interactive competence. 
There is no universal optimum between realism and control. 
Benchmarks should instead state which deployment conditions they model faithfully, which they deliberately abstract away, and which claims their level of fidelity can and cannot support.

\vspace{-4pt}
\paragraph{Evaluator and Counterpart Dependence.}
As user simulators, model judges, human experts, and counterpart agents become standardized evaluation infrastructure, they begin to shape what counts as successful interaction. 
Scores may then reward adaptation to particular evaluators or counterpart policies rather than the intended capability. 
This creates a construct-validity risk: systems may perform well under one judge, simulator, or expert group but fail under plausible alternatives. 
Future benchmarks should test whether conclusions remain stable across evaluator and counterpart variants.



\section{Scope and Implications of the Position}


The implementation limitations and risks discussed above address several broad alternative views: that interactive evaluation may be costly, protocol-dependent, difficult to reproduce, sensitive to simulator artifacts, or vulnerable to trajectory-level gaming. 
Additional alternative views are discussed in Appendix~\ref{sec:alternative-views}. 
Our position is therefore not that every benchmark should become interactive, nor that interactive evaluation always requires expensive high-fidelity simulation. 
Instead, the central question is whether interaction is constitutive of the capability claim, and what evaluative claims trajectory evidence can legitimately support. 
This section clarifies that scope.

\paragraph{Interactive Evaluation Is Not Just Agent Evaluation.}

Our argument is not that every ``agent`` requires interactive evaluation. 
Agent benchmarks can remain response-centered evaluation when actions do not affect later conditions or only the final output is judged. 
Conversely, systems that are not described as autonomous agents may still require interactive evaluation if their behavior unfolds through tools, web environments, users, or other external loops. 
The key question is therefore not what the system is called, but what evidence the evaluation needs in order to support its claim.

\paragraph{Trajectory-Level Evaluation Is Claim-dependent.}

We do not attempt to claim that every evaluation should become a high-fidelity simulation at the cost of scalability, or that task success should be replaced. Task success remains indispensable when the intended claim concerns final completion. The problem arises when a benchmark records trajectories but treats them only as evidence for a final success label. If the claim concerns process quality, recoverability, safety, efficiency, coordination, or robustness, the evaluation program must preserve and score the relevant trajectory evidence. 
The design challenge is to match the cost and fidelity of interactive evaluation to the claim.
\vspace{-2mm}
\section{Conclusion}
\vspace{-2mm}

In this position paper, we argue that interactive evaluation must be designed, not merely adopted. 
As AI systems increasingly act through consequential interactions, the field needs a systematic and unified framework for designing interactive evaluations that support comparison, reproducibility, and extension. We frame interactive evaluation as trajectory-based, system-level evaluation under action-dependent conditions, organized by two questions: what interaction artifacts enter evaluation, and how those artifacts are mapped to judgments. 
This framing clarifies why response-centered benchmarks remain useful but insufficient, why current interactive benchmarks should not be treated as interchangeable, and what the field must build next such as explicit protocols, richer trajectory measures, robustness tests, shared infrastructure, and reporting standards that make interactive scores interpretable. We therefore call on the community to design interactive evaluation before merely adopting it as the next benchmark format.

\section*{Acknowledgment}

We thank Daniel Fried (CMU) and Katie M. Collins (Cambridge/MIT/Princeton) for discussions and helpful feedback on earlier versions of this paper.

\clearpage
\newpage
\bibliographystyle{assets/plainnat}
\bibliography{main}

\clearpage
\newpage
\beginappendix

\section{More Alternative Views}
\label{sec:alternative-views}

This paper argues that interactive evaluation requires a design science for evaluating systems acting through trajectories. 
As we clarified throughout our main paper, this position does not imply that all evaluations should become interactive, that response-centered benchmarks should be discarded, or that every recorded trajectory should be converted into a single process score. 
Below we discuss several viable alternative views and clarify how our position accommodates or responds to them.

\textbf{Alternative view 1: Interactive evaluation is not distinct from dynamic, live, executable, or
holistic evaluation.}

One might argue that the field already has many extensions of traditional benchmarking: live benchmarks, contamination-resistant benchmarks, dynamic datasets, executable coding tasks, tool-use benchmarks, holistic evaluation suites, and continuously refreshed evaluations. 
From this view, interactive evaluation may appear to rename existing trends rather than identify a distinct
evaluation paradigm.

\textbf{Response.}
These efforts are closely related and often complementary, but they are not identical to interactive evaluation as defined in this paper. 
The distinction is not whether a benchmark is difficult, fresh, realistic, executable, or multi-step. 
The distinction is whether the admissible evidence includes trajectories generated by consequential interaction, and whether the evaluation program maps those trajectories to judgments about system-level performance.

A live benchmark can remain response-centered if it simply refreshes fixed instances and scores final answers. 
An executable benchmark can remain response-centered if execution is used only to check a final submitted artifact. 
A tool-use benchmark may be interactive if tool results change subsequent observations and the trajectory matters to the judgment, but tool use alone is not sufficient. 
Likewise, a holistic evaluation suite may cover many capabilities without evaluating action-dependent trajectories. 
Interactive evaluation is therefore defined by action dependence and trajectory-to-judgment mapping, not by novelty, realism, or breadth.

This boundary matters because different benchmark families support different claims. 
Dynamic and live benchmarks address contamination and temporal freshness. 
Executable benchmarks improve verifiability. Holistic suites broaden coverage. 
Interactive evaluation addresses a different question: how should we evaluate systems whose behavior unfolds through state, feedback, tools, users, or other agents, such that earlier actions shape later evidence and the meaning of success? 
Our position is that this question requires explicit design principles rather than being absorbed into existing benchmark categories.

\textbf{Alternative view 2: Interactive evaluation is too expensive and should be used sparingly.}

A reasonable objection is that interactive evaluation can increase the cost of benchmarking. 
Long trajectories, executable environments, state snapshots, user simulators, counterpart agents, repeated runs, and trajectory-level judging can all make evaluation harder to run and harder to audit. 
If interactive evaluation becomes the default benchmark format, it may privilege large labs and reduce the accessibility that made response-centered benchmarks scientifically valuable.

\textbf{Response.} We agree that interactive evaluation should not be used for everything. 
The right criterion is not whether an evaluation can be made interactive, but whether the capability claim requires interaction as evidence. 
If the relevant evidence is already contained in a fixed input and final output, then a response-centered evaluation may be the better design: cheaper, cleaner, and easier to reproduce.
Our position is instead claim-dependent. 
Interactive evaluation is needed when earlier system behavior changes later observations, states, opportunities, costs, risks, or counterpart responses, and when those changes matter to the judgment being made.

At the same time, the cost objection should not be overstated. 
Many apparently response-centered evaluations already require intermediate artifacts: generated code, tool calls, retrieved documents, execution traces, plans, test logs, edited files, or state changes. 
When such artifacts are already produced as part of evaluation, trajectory-level evaluation may not require a wholly new evaluation regime. 
It may instead require using evidence that the benchmark already creates but currently throws away. 
In these cases, interactive evaluation can improve evidential efficiency: the same evaluation run can support not only a final success judgment, but also claims about recovery, constraint satisfaction, action economy, unsafe operations, or robustness. 
The design challenge is therefore to add trajectory-level judgments only where they provide additional validity relative to their cost.

\textbf{Alternative view 3: Response-centered evaluation remains sufficient for most purposes.}

Another view is that response-centered evaluation has been successful because it captures the most important comparison signal. 
Fixed inputs, standardized outputs, and simple scoring support clear leaderboards, scalable experimentation, and cumulative progress. 
From this perspective, interactive evaluation risks replacing a robust paradigm with a more complex one whose benefits are uncertain.

\textbf{Response.} We do not argue that response-centered evaluation should be discarded. 
On the contrary, it remains indispensable when the object of evaluation is naturally an input-output mapping, when final correctness is the relevant claim, or when interaction is incidental rather than consequential. 
The boundary is evaluative, not stylistic. 
Multiple turns, tool calls, or generated intermediate text do not by themselves require interactive evaluation. 
They require interactive evaluation only when the trajectory changes what evidence is available and what judgment the score can support.

The limitation of response-centered evaluation appears when final responses are no longer sufficient evidence. 
A system that reaches the correct final state by corrupting persistent state, ignoring user constraints, manipulating a counterpart, overusing tools, or failing to recover from induced errors has not demonstrated the same capability as a system that reaches the same outcome safely and robustly.
In such cases, final success remains important, but it is incomplete. 
Interactive evaluation supplements response-centered evaluation exactly where process, state, feedback, and consequence change the meaning of success.

\textbf{Alternative view 4: Task success is the only robust metric; process metrics are subjective.}

A strong objection is that final task success has clearer face validity than process-level measurement.
By contrast, metrics for process quality, communication, risk, recovery, or efficiency may encode contestable assumptions. 
If benchmarks assign scores to trajectories, they may reduce comparability or give a false sense of objectivity to normative choices about what counts as a good process.

\textbf{Response.} This objection identifies a real danger, but it supports our design-science view rather than undermining it. 
The problem is not that process metrics involve choices; all evaluation metrics involve choices.
The problem is when those choices are hidden. 
Outcome-only metrics can appear neutral while silently permitting unsafe, brittle, manipulative, or wasteful behavior whenever such behavior still achieves the final goal. 
A benchmark that scores only success is not free of process assumptions; it implicitly assumes that all successful trajectories are equivalent for the intended claim.

The appropriate response is not to collapse outcome, process, and risk into a single opaque score.
Interactive evaluations should report these dimensions separately whenever they support distinct claims. 
Final success should remain visible. 
Process measures should be tied, where possible, to externally checkable trajectory properties: state changes, action counts, constraint violations, failed tool calls, recovery after perturbation, rollback behavior, unsafe operations, communication failures, or robustness across seeds and counterpart variants. 
Aggregate scores may still be useful for ranking, but they should be treated as summaries of reported dimensions, not as substitutes for them. 
This is why interactive evaluation requires explicit evaluation programs rather than ad hoc trajectory judgment.

\textbf{Alternative view 5: Trajectory-level evaluation may reward performative behavior rather
than capability.}

If benchmarks judge trajectories, systems may learn to produce trajectories that look good to humans or model judges. 
They may generate verbose plans, artificial self-corrections, or judge-pleasing rationales without improving actual competence. 
In the worst case, trajectory scoring could become another surface to game.

\textbf{Response.} This is an important trajectory-level analogue of benchmark gaming. 
It is one reason we distinguish trajectory evidence from exposed reasoning style. 
Interactive evaluation should not reward a trajectory because it appears thoughtful, verbose, or human-like. 
It should reward trajectory properties that are relevant to the capability claim and, when possible, grounded in observable interaction outcomes.

For example, recovery should be tested by whether a system detects and repairs an actual induced failure, not by whether it says that it is being careful. 
Efficiency should be measured by action cost, redundant tool use, or unnecessary state changes, not by whether the transcript appears concise.
Safety should be measured through constraint violations, unsafe operations, or harmful state changes, not merely by safety-themed language. 
This shifts trajectory evaluation away from stylistic judgment and toward behavioral evidence. 
It also reinforces the need for perturbations, audits of suspiciously efficient trajectories, replayable logs, and transparent reporting of what the evaluator actually scores.

\textbf{Alternative view 6: Interactive evaluation conflates model capability with system engineering.}

A further objection is that interactive benchmarks often evaluate more than the base model. Tool wrappers, memory, retrieval systems, planners, sandboxes, interface affordances, orchestration policies, and prompting strategies may dominate performance. 
If so, interactive evaluation may make it difficult to know whether progress comes from better models or better systems engineering.

\textbf{Response.} We agree that interactive evaluation often evaluates systems rather than isolated models. 
This is not a defect of the paradigm; it is a property of the deployment settings that motivate it. 
When an AI system acts through tools, environments, users, memory, or other agents, the relevant object of evaluation is frequently the assembled system. 
A benchmark that ignores wrappers, permissions, state, orchestration, or tool interfaces may provide a cleaner model-level comparison, but it may not support claims about deployed interactive behavior.

The implication is that benchmark reports must distinguish model-level and system-level claims.
If the goal is to compare base models, then the surrounding scaffold should be controlled and reported. 
If the goal is to compare complete agents or deployed assistants, then the scaffold is part of the evaluated object and should be documented as such. 
Interactive evaluation therefore raises the standard for reporting: model identity alone is insufficient. 
Evaluations should specify tools, memory, retrieval, prompts, orchestration, sandbox permissions, environment versions, and logging protocols so that readers can interpret what the score is actually evidence about.

\section{Related Works}

As is the nature of a position paper, we discuss the most closely related benchmarks and evaluation paradigms in the main text when defining interactive evaluation and constructing the taxonomy. 
This appendix notes adjacent areas that inform the position but are not central to the main argument.

\paragraph{Detection \& Mitigation of Response-Centered Evaluation Issues.}

A large body of prior work studies limitations of response-centered evaluation, including benchmark leakage~\citep{Jacovi2023StopUT, Balloccu2024LeakCR}, data contamination~\citep{Sainz2023NLPEI, Xu2024BenchmarkDC}, benchmark overfitting~\citep{McCoy2019RightFT}, brittleness under distribution shift~\citep{Ailem2024ExaminingTR, han2024context}, and the use of private, hidden, or live evaluation to reduce memorization and gaming~\citep{Kiela2021DynabenchRB, White2024LiveBenchAC}. To address these limitations, researchers have proposed contamination audits~\citep{Golchin2023TimeTI, Deng2023InvestigatingDC}, benchmark decontamination~\citep{Zhu2024InferenceTimeDR}, adversarial or counterfactually constructed examples~\citep{Ribeiro2020BeyondAB, Kiela2021DynabenchRB}, dataset refreshment~\citep{Ying2024AutomatingDU}, live evaluation~\citep{Chandran2024PrivateBT}, more explicit reporting of evaluation conditions~\citep{Liang2023HolisticEO, Jacovi2023StopUT}, and meta-annotation schemes to understand model performance across multiple benchmarks~\citep{zhou2026general}. 
These efforts have been essential for clarifying when fixed-instance benchmark scores provide reliable evidence of model capability. 
These concerns motivate our risk analysis, but our focus is different: Here we ask how these problems change when the evaluated system acts through consequential trajectories rather than producing isolated responses, and what additional design requirements this creates for interactive evaluation.

\paragraph{Other Extensions to Traditional Benchmarks.}

Traditional benchmarks have also been extended in several directions beyond the response-centered format. 
Live and continuously refreshed benchmarks update evaluation instances over time to reduce memorization and better reflect current model capabilities~\citep{White2024LiveBenchAC, Jain2024LiveCodeBenchHA, Kasai2022RealTimeQW}. 
Dynamic and procedurally generated benchmarks vary tasks, constraints, or environments to reduce benchmark-specific overfitting~\citep{Zhu2023DyValDE, Qian2024VarBenchRL, ying2026ai}. 
Private or hidden benchmarks restrict access to test instances to limit leakage and gaming~\citep{Oren2023ProvingTS, Zhao2024MMLUCFAC}. 
Other benchmarks add execution, tools, browsing, long-context inputs, or real-world task constraints so that model outputs can be checked against tests, external evidence, or richer task specifications.
These efforts are complementary, but they mainly improve how benchmark instances are sourced, protected, refreshed, or verified. 
Our focus is action-dependent evaluation: settings where earlier system behavior shapes later observations, states, opportunities, or counterpart responses. 
The central question is therefore not only how to make tasks fresher or harder, but how trajectory evidence should be mapped to claims about system-level performance.


\section{Additional Details for the Roadmap}
\label{app:benchmark_collection}

\subsection{Roadmap Categories.}
\label{app:roadmap_category}
To examine the temporal pattern behind Figure~\ref{fig:roadmap}, we use three roadmap categories. 
\textbf{Response-centered evaluation} refers to benchmarks where the main evidence is a final response to a fixed instance. 
\textbf{Task-driven extensions} refer to benchmarks that add execution, tools, web access, or multi-step task completion, but remain primarily organized around fixed task outcomes rather than consequential trajectory evidence. 
\textbf{Interactive evaluation} refers to benchmarks where trajectories generated by consequential interaction enter the admissible evidence and are mapped to judgments about system-level performance.

\subsection{Benchmark Collection.}
\label{app:collection}

To analyze temporal trends across the three roadmap categories, the representative benchmark list alone is not sufficient. We therefore expand it through two semi-automated retrieval channels. First, we perform citation-based snowball sampling from the representative benchmarks and retain citing papers whose titles match benchmark-related keywords, such as ``bench,'' ``arena,'' and ``gym.'' Second, snowball sampling can miss early-year work that predates or coincides with our anchors. 
We therefore run stage-neutral Semantic Scholar searches for each year from 2020 to 2026.

We then deduplicate papers across channels using arXiv IDs when available and normalized titles otherwise, and apply a shared quality filter to obtain the final candidate set. A paper is retained if it appears in a top venue, or has citation velocity at least 1.5, or has at least 50 GitHub stars. We define citation velocity as
\[
\mathrm{CitationVelocity}(p)
=
\frac{\mathrm{Citations}(p)}
{\max\!\left(\mathrm{MonthsSincePublication}(p),\, 3\right)} .
\]

All candidate papers are then classified by an LLM-based classifier into the three roadmap categories or \emph{Not Relevant}; only papers assigned to the three roadmap categories are included in the trend analysis. 
Classification uses each paper's title and abstract, follows the roadmap definitions above, and is based on the paper's primary contribution rather than whether it merely evaluates a new model on existing benchmarks. 
We validate the classifier on the manually curated anchor set and apply it to the expanded corpus only after it achieves over 90\% agreement with our manual labels. 
For ambiguous cases, the classifier assigns Interactive Evaluation only when the paper explicitly emphasizes trajectory evidence, stateful interaction, or agent feedback loops. 
We use this analysis as descriptive evidence for broad temporal trends, rather than as an exhaustive census of all evaluation work.



\subsection{Industry-Academic Comparison.}
Panel (c) of Figure~\ref{fig:roadmap} compares evaluation-stage composition between recent frontier industry reports and academic benchmark papers from 2024--2026. 
The industry sample contains 43 distinct benchmark families extracted from the most recent public model cards or technical reports of OpenAI, Anthropic, Google DeepMind, and Alibaba/Qwen, with each benchmark family counted once per source document regardless of variants or subtasks. 
The academic sample is the 2024--2026 subset of the benchmark collection described above, containing 360 benchmark papers.
Bars report percentage shares within each group, so each group sums to 100\%. 
A Pearson \(\chi^2\) test gives \(\chi^2(2)=7.09, p=0.029\), indicating a statistically significant difference in stage distribution. 
We interpret this comparison as descriptive evidence that the transition toward task-driven and interactive evaluation is uneven across the evaluation ecosystem, rather than as an exhaustive census of either community.

\section{Illustrative Scenarios}
\label{app:illustrative_scenario}

This section provides two illustrative scenarios showing how the proposed framework can be applied end to end. 
The goal is to make concrete how interactive evaluation changes both sides of the evaluation mapping \(E:\mathcal{X}\rightarrow \mathcal{Y}\): what trajectory evidence enters \(\mathcal{X}\), and what evaluation program \(E\) is needed to turn that evidence into valid claims.

We choose these two scenarios because they represent two broad classes of settings where interactive evaluation is especially valuable. 
Coding-agent evaluation illustrates domains involving multi-step reasoning, iterative planning, execution, debugging, and revision, where the trajectory itself contains important evidence about process quality, recoverability, and robustness beyond final task completion. 
Multi-agent social evaluation illustrates settings in which simulation, interaction, and adaptive environments are necessary for probing coordination, negotiation, communication, and other socially situated behaviors that cannot be captured through isolated responses alone. 
While the examples are concrete, the underlying design principles generalize more broadly across many other interactive evaluation settings within each category.

\subsection{Coding Agents.}

\paragraph{Mapping to the Definition.} Coding-agent evaluation naturally benefits from interactive evaluation as the system acting in a repository-level feedback loop. 
The agent works through repository-level actions: inspecting files, running commands, observing test failures or error traces, editing code, and revising its solution based on feedback~\citep{Yang2024SWEagentAI,kumarappan2025leanagent,wang2025openhands}. 
Because these actions shape later observations and repair opportunities, the relevant evidence is the full interaction trajectory instead of single final patch.

\paragraph{From Response-Centered to Interactive Evaluation.} Traditional code-generation benchmarks evaluate a final output against references or tests~\citep{Hendrycks2021MeasuringCC, Austin2021ProgramSW}, but repository-level agents raise broader claims. Two agents may both pass hidden tests, yet one may rely on principled diagnosis and localized edits while another relies on brittle edits or visible-test overfitting. A final pass/fail label treats these trajectories as equivalent, even though they support different claims about debugging competence, maintainability, and deployment reliability.

\paragraph{Mapping to the Taxonomy.} In our taxonomy, coding-agent evaluation mainly belongs to Tools and Environments because the agent interacts with repositories, command-line tools, test suites, and executable environments. The input artifact \(\mathcal{X}\) should include repository state, issue text, tool calls, file edits, test executions, error traces, and final patch. On the evaluation-program axis, coding agents should require not only Task Success, but also Process Quality and Efficiency, Recoverability and Robustness, and risk-sensitive evaluation for transparent failure diagnosis.

\paragraph{Applying the Design Principles.} The evaluation program \(E\) should distinguish whether the issue is resolved, whether the patch is localized and maintainable~\citep{Zhu2026NeedleIT}, whether the agent uses tests and errors to recover, and whether it avoids collateral damage. SWE-bench and recent coding-agent benchmarks already evaluate agents in repository-level or long-horizon programming settings~\citep{jimenez2023swe, khatua2026cooperbench, feng2026longcli}. However, they still often reduce repository-level interaction to final resolution, leaving diagnosis quality, recovery behavior, patch locality, and collateral risk under-specified. Concretely, coding-agent benchmarks should specify tool access, test access, retry policy, repository reset conditions, and logging format, and should report final resolution separately from trajectory-level measures.

\paragraph{Risks and Open Issues.}Coding-agent evaluation also exposes trajectory-level risks: agents may game visible tests, exploit benchmark-specific repository patterns, or produce patches that pass current tests while introducing hidden regressions. Scores can also be sensitive to environment setup, dependency versions, tool access, timeout limits, and retry policies. Future evaluations therefore need replayable trajectory logs, environment versioning, and reporting standards that distinguish genuine debugging competence from benchmark-specific exploitation.

\subsection{Multi-Agent Social Systems.}

\paragraph{Mapping to the Definition.}
Multi-agent social evaluation is an interactive setting because the system acts in a social feedback loop with other agents whose beliefs, strategies, and future behavior can change in response to its actions~\citep{Davidson2024EvaluatingLM, Abdelnabi2023CooperationCA}. 
The evaluated agent does not merely produce an isolated utterance; it communicates, negotiates, coordinates, refuses, adapts to counterpart behavior, and may adjust its strategy as the interaction unfolds. 
These actions shape subsequent messages, commitments, conflicts, and opportunities for coordination, so the relevant evaluation evidence is the full social trajectory rather than only the final group outcome.

\paragraph{From Response-Centered to Interactive Evaluation.}
Response-centered evaluation can assess an isolated utterance, but social agents raise broader claims about coordination, fairness, communication, and robustness. For example, a negotiation agent may make a reasonable and clearly phrased offer in one turn, while still exploiting counterpart concessions, treating stronger and weaker counterparts differently, or breaking down under unusual communication styles. 
These properties require evidence from the full interaction trajectory, not only from an isolated response.

\paragraph{Mapping to the Taxonomy.}
In our taxonomy, multi-agent social evaluation mainly belongs to Other Agents because the system interacts with counterparts whose goals, information, and behavior may change over time. 
The input artifact \(\mathcal{X}\) should include messages, role assignments, private and shared information, commitments, proposals, refusals, conflict points, counterpart behavior, and final outcomes. 
On the evaluation-program axis, social agents require not only Task Success, but also Safety, Alignment, and Social Competence, as well as Recoverability and Robustness when agents must repair misunderstanding or adapt to strategic counterparts.

\paragraph{Applying the Design Principles.}
The evaluation program \(E\) should distinguish whether the social goal is completed, whether agents coordinate effectively, or whether communication is fair and transparent.
Benchmarks such as SOTOPIA, MultiAgentBench, BattleAgentBench, MASEval, and CooperBench already evaluate agents in cooperative, competitive, or mixed-motive social settings~\citep{zhou2023sotopia, zhu2025multiagentbench, wang2024battleagentbench, emde2026maseval, khatua2026cooperbench}. 
However, they still often reduce social interaction to aggregate success or judge-level preferences, leaving coordination process, fairness, recovery from misunderstanding, and social risk under-specified~\citep{xuan2026socialveil}. 
Concretely, multi-agent benchmarks should specify role assignments, information asymmetries, counterpart behavior, memory, turn-taking rules, stopping criteria, and hidden goals or constraints, and should report group success separately from trajectory-level measures.

\paragraph{Risks and Open Issues.}
Multi-agent social evaluation introduces interaction-specific risks. 
An agent may appear cooperative with one counterpart but become exploitative, evasive, or overly deferential with another; small changes in roles, private information, or power asymmetry can change what competence means~\citep{Lewis2017DealON}. 
Because judgments depend on norms, incentives, and counterpart behavior, scores may conflate genuine coordination with persuasion, pressure, or strategic withholding~\citep{Manheim2018CategorizingVO,Perez2022DiscoveringLM}. 
Evaluations should make roles, incentives, norms, and judge criteria explicit, and report whether success reflects robust social competence or undesirable social strategies.


\section{Representative Benchmark List}
\label{app:benchmark_list}

This section reports the representative benchmark list (Fig.~\ref{fig:roadmap}) and the taxonomy discussion (Fig.~\ref{fig:taxonomy}) in the main paper. 
The list is intended as a transparent resource for the roadmap analysis rather than an exhaustive census of all evaluation benchmarks. 
We record metadata such as citation counts and GitHub stars as accurately as possible at the time of collection\footnote{The data in this table were last updated on May 7, 2026.}. 
However, note that these values should be interpreted only as approximate indicators rather than definitive measures of benchmark importance or influence.
In few cases, GitHub star counts may be unavailable for older benchmarks or for works without an official public repository, in which case we leave the field blank rather than infer a value. 
Some other benchmarks created Github repositories substantially later than their release dates, which may also make star counts unevenly comparable across works.
Similarly, citation counts can reflect many factors beyond evaluative significance, including publication venue, age, community size, and research trends.
We therefore include these metrics primarily as suggestive signals that help visualize the current evaluation landscape, while aiming to make our benchmark selection and categorization process easier to inspect, reproduce, and refine in future work, rather than presenting them as authoritative rankings of benchmark quality or impact.

\vspace{1em}
 
\begin{longtable}{@{}p{6cm}cp{6cm}rr@{}}
\toprule
\textbf{Benchmark} & \textbf{Year} & \textbf{Task Type} & \textbf{Citations} & \textbf{Stars} \\
\midrule
\endfirsthead
 
\multicolumn{5}{l}{\small\textit{(continued from previous page)}} \\
\toprule
\textbf{Benchmark} & \textbf{Year} & \textbf{Task Type} & \textbf{Citations} & \textbf{Stars} \\
\midrule
\endhead
 
\midrule
\multicolumn{5}{r}{\small\textit{(continued on next page)}} \\
\endfoot
 
\bottomrule
\endlastfoot
 
\multicolumn{5}{l}{\cellcolor{rccolor}\textbf{\textcolor{rcheader}{Stage 1: Response-Centered Evaluation}}}\\
\midrule
  \href{https://arxiv.org/abs/1606.05250}{SQuAD}~\citep{rajpurkar2016squad} & 2016 & Reading Comprehension & 11,679 & --- \\
  \href{https://arxiv.org/abs/1804.07461}{GLUE}~\citep{wang2019gluemultitaskbenchmarkanalysis} & 2018 & Reading Comprehension & 10,589 & --- \\
  \href{https://arxiv.org/abs/1903.00161}{DROP}~\citep{dua2019dropreadingcomprehensionbenchmark} & 2019 & Reading Comprehension & 1,438 & --- \\
  \href{https://arxiv.org/abs/1811.00937}{CommonsenseQA}~\citep{talmor2019commonsenseqa} & 2019 & Commonsense Reasoning & 2,666 & 168 \\
  \href{https://arxiv.org/abs/2009.03300}{MMLU}~\citep{hendrycks2020measuring} & 2020 & Knowledge \& Multitask Reasoning & 7,833 & 1.4k\\
  \href{https://arxiv.org/abs/2110.14168}{GSM8k}~\citep{cobbe2021training} & 2021 & Math Reasoning & 8,894 & 1.4k \\
  \href{https://arxiv.org/abs/2103.03874}{MATH}~\citep{hendrycks2021measuringmathematicalproblemsolving} & 2021 & Math Reasoning & 397 & 433 \\
  \href{https://arxiv.org/abs/2109.00110}{MiniF2F}~\citep{zheng2021minif2f} & 2021 & Formal Theorem Proving & 5,004 & 1.3k \\
  \href{https://arxiv.org/abs/2108.07732}{MBPP}~\citep{Austin2021ProgramSW} & 2021 & Code Generation & 3,556 & 37.8k \\
  \href{https://arxiv.org/abs/2107.03374}{HumanEval}~\citep{chen2021evaluating} & 2021 & Code Generation & 9,594 & 3.2k \\
  \href{https://arxiv.org/abs/2206.04615}{Big-bench}~\citep{srivastava2023beyond} & 2022 & Broad Capability Probing & 2,560 & 3.2k \\
  \href{https://arxiv.org/abs/2109.07958}{TruthfulQA}~\citep{lin2022truthfulqa} & 2022 & Truthfulness \& Factuality & 3,638 & 910 \\
  \href{https://arxiv.org/abs/2306.15626}{LeanDojo}~\citep{yang2023leandojo} & 2023 & Formal Theorem Proving & 592 & 796 \\
  \href{https://arxiv.org/abs/2306.05685}{MT-Bench}~\citep{zheng2023judging} & 2023 & Human Preference Evaluation & 9,124 & 39.5k \\
  \href{https://arxiv.org/abs/2308.14508}{LongBench}~\citep{bai2024longbench} & 2023 & Long-Context Understanding & 1,191 & 1.1k \\
  \href{https://arxiv.org/abs/2403.04132}{Chatbot Arena}~\citep{chiang2024chatbot} & 2024 & Human Preference Evaluation & 1,196 & 39.5k \\
  \href{https://arxiv.org/abs/2402.17753}{LoCoMo}~\citep{maharana2024evaluatinglongtermconversationalmemory} & 2024 & Long-term Memory & 390 & 828 \\
  \href{https://arxiv.org/abs/2404.04475}{AlpacaEval}~\citep{dubois2024length} & 2024 & Human Preference Evaluation & 1,048 & 2k \\
  \href{https://arxiv.org/abs/2410.07985}{Omni-Math}~\citep{gao2025omni} & 2024 & Math Reasoning & 175 & 93 \\
  \href{https://arxiv.org/abs/2410.10813}{LongMemEval}~\citep{wu2024longmemeval} & 2025 & Long-term Memory & 234 & 738 \\
 
\multicolumn{5}{l}{\cellcolor{tdcolor}\textbf{\textcolor{tdheader}{Stage 2: Task-Driven Extensions}}}\\
\midrule
  \href{https://arxiv.org/abs/2310.06770}{SWE-bench}~\citep{jimenez2023swe} & 2023 & Code \& Software Engineering & 2,397 & 4.8k \\
  \href{https://arxiv.org/abs/2304.08244}{API-Bank}~\citep{li2023api} & 2023 & Tool Use \& API Calling & 502 & 1.6k \\
  \href{https://arxiv.org/abs/2306.06070}{Mind2Web}~\citep{deng2023mind2web} & 2023 & Web Navigation & 1,113 & 986 \\
  \href{https://arxiv.org/abs/2311.12983}{GAIA}~\citep{mialon2023gaia} & 2023 & Tool Use \& API Calling & 829 & --- \\
  \href{https://arxiv.org/abs/2307.16789}{ToolBench}~\citep{qin2023toolllm} & 2023 & Tool Use \& API Calling & 1,451 & 5.6k \\
  \href{https://arxiv.org/abs/2311.18760}{TaskBench}~\citep{shen2024taskbench} & 2023 & Task Automation \& Planning & 142 & 24.7k \\
  \href{https://arxiv.org/abs/2403.07974}{LiveCodeBench}~\citep{Jain2024LiveCodeBenchHA} & 2024 & Code Generation \& Execution & 1,384 & 857 \\
  \href{https://arxiv.org/abs/2403.07714}{Stabletoolbench}~\citep{guo2024stabletoolbench} & 2024 & Tool Use \& API Calling & 178 & 232 \\
  \href{https://arxiv.org/abs/2402.01622}{TravelPlanner}~\citep{xie2024travelplanner} & 2024 & Planning \& Constraint Satisfaction & 372 & 510 \\
  \href{https://arxiv.org/abs/2505.12331v1}{OSS-Bench}~\citep{jiang2025oss} & 2025 & Code \& Software Engineering & 3 & 8 \\
  \href{https://arxiv.org/abs/2508.13186}{MM-BrowseComp}~\citep{li2025mm} & 2025 & Web Navigation & 21 & 67 \\
  \href{https://arxiv.org/abs/2504.12516}{BrowseComp}~\citep{wei2025browsecomp} & 2025 & Web Navigation & 364 & 4.4k \\
  \href{https://arxiv.org/abs/2601.18137}{DeepPlanning}~\citep{zhang2026deepplanning} & 2026 & Planning \& Constraint Satisfaction & 10 & 16.3k \\
  \href{https://arxiv.org/abs/2601.11868}{Terminal-Bench}~\citep{merrill2026terminal} & 2026 & Code \& Software Engineering & 80 & 2.2k \\
  \href{https://arxiv.org/abs/2602.14337}{Longcli-bench}~\citep{feng2026longcli} & 2026 & Code \& Software Engineering & 5 & 33 \\
 
\multicolumn{5}{l}{\cellcolor{iecolor}\textbf{\textcolor{ieheader}{Stage 3: Interactive Evaluation}}}\\
\midrule
  \href{https://arxiv.org/abs/2407.18901}{AppWorld}~\citep{trivedi2024appworld} & 2024 & App / Software Use & 146 & 413 \\
  \href{https://arxiv.org/abs/2405.14573}{AndroidWorld}~\citep{rawles2024androidworld} & 2024 & Web / Computer Use & 267 & 752 \\
  \href{https://arxiv.org/pdf/2406.12045}{$\tau$-bench}~\citep{yao2024tau} & 2024 & Tool / Assistant & 496 & 1.2k \\
  \href{https://arxiv.org/abs/2403.07718}{WorkArena}~\citep{drouin2024workarena} & 2024 & Web / Computer Use & 229 & 1.2k \\
  \href{https://arxiv.org/abs/2309.15817}{ToolEmu}~\citep{ruan2024identifying} & 2024 & Tool / Assistant & 364 & 202 \\
  \href{https://arxiv.org/abs/2401.13649}{VisualWebArena}~\citep{koh2024visualwebarena} & 2024 & Web / Computer Use & 527 & 466 \\
  \href{https://arxiv.org/abs/2405.07960}{AgentClinic}~\citep{schmidgall2024agentclinic} & 2024 & Tool / Assistant  & 221 & 305 \\
  \href{https://arxiv.org/abs/2404.07972}{OSWorld}~\citep{xie2024osworld} & 2024 & Web / Computer Use & 616 & 2.8k \\
  \href{https://arxiv.org/abs/2406.13352}{AgentDojo}~\citep{debenedetti2024agentdojo} & 2024 & Web / Computer Use & 561 & 2.8k \\
  \href{https://arxiv.org/abs/2307.13854}{WebArena}~\citep{zhou2023webarena} & 2024 & Web / Computer Use & 1,174 & 1.5k \\
  \href{https://arxiv.org/abs/2310.11667}{Sotopia}~\citep{zhou2023sotopia} & 2024 & Social Interaction & 391 & 300 \\
  \href{https://arxiv.org/abs/2507.22034}{UserBench}~\citep{qian2025userbench} & 2025 & Tool / Assistant & 33 & 60 \\
  \href{https://arxiv.org/pdf/2501.11067}{Intellagent}~\citep{levi2025intellagent} & 2025 & Social Interaction & 18 & 1.2k \\
  \href{https://arxiv.org/abs/2412.14470}{Agent-SafetyBench}~\citep{zhang2024agent} & 2025 & Tool / Assistant & 157 & 134 \\
  \href{https://arxiv.org/abs/2408.04682}{ToolSandbox}~\citep{lu2025toolsandbox} & 2025 & Tool / Assistant & 130 & 247 \\
  \href{https://arxiv.org/abs/2503.01935}{Multi-agent Bench}~\citep{zhu2025multiagentbench} & 2025 & Social Interaction & 152 & 46 \\
  \href{https://arxiv.org/pdf/2502.15840}{Vending-Bench}~\citep{backlund2025vending} & 2025 & Social Interaction & 34 & 3.3k \\
  \href{https://arxiv.org/abs/2512.01078}{SimWorld}~\citep{ren2025simworld} & 2025 & Embodied / Open-World & 3 & 561 \\
  \href{https://arxiv.org/abs/2509.17158}{ARE (GAIA2)}~\citep{froger2025scaling} & 2025 & App / Software Use & 16 & 482 \\
  \href{https://arxiv.org/abs/2504.10445}{RealWebAssist}~\citep{ye2026realwebassist} & 2025 & Web / Computer Use & 17 & 10 \\
  \href{https://arxiv.org/abs/2601.13295}{CooperBench}~\citep{khatua2026cooperbench} & 2026 & Web / Social Interaction  & 4 & 13 \\
  \href{https://arxiv.org/abs/2510.06288v3}{BuilderBench}~\citep{ghugare2025builderbench} & 2026 & Embodied / Open-World & 2 & 32 \\
 
\end{longtable}

\end{document}